# Modified Segmentation Algorithm for Recognition of Older Ge'ez Scripts Written on Vellum

**Girma Negashe[1]**

[1]Faculty of Informatics, University of Gondar
Gondar, Ethiopia
e-mail: girma.sisay@uog.edu.et

**Adane Mamuye[2]**

[2]Faculty of Informatics, University of Gondar
Gondar, Ethiopia
e-mail: adane.letta@uog.edu.et

**Abstract:** Recognition of handwritten document aims at transforming document images into a machine understandable format. Handwritten document recognition is the most challenging area in the field of pattern recognition. It becomes more complex when a document was written on a vellum before hundreds of years, like older Ge'ez scripts. In this study, we introduced a modified segmentation approach to recognize older Ge'ez scripts. We used adaptive filtering for noise reduction, Isodata iterative global thresholding for document image binarization, modified bounding box projection to segment distinct strokes between Ge'ez characters, numbers and punctuation marks. SVM multiclass classifier scored 79.32% recognition accuracy with the modified segmentation algorithm.

**Keywords-** *OCR, Multiclass classification, Modified character segmentation, Ge'ez scripts*

## 1. INTRODUCTION

Optical Character Recognition (OCR) translates images of printed, typewritten or handwritten characters into machine editable format. The development, OCR for handwriting characters is an interesting area of pattern recognition [1] that involves a number of stages, namely preprocessing, segmentation, feature extraction, classification and actual recognition [2].

These days OCR systems can read a variety of documents written in several languages such as English, Latin, Japanese, Chinese, Hindu, Arabic, Swedish and Russian [2]. However, recognition of Ge'ez scripts, one of the ancient scripts and the liturgical language of Ethiopia, written on the vellum are at an early stage.

Older handwritten Ge'ez scripts are different from modern documents in various ways, such as writing style, morphological structure and writing materials [3]. Figure 1, shows the Ge'ez scripts with degraded quality written on the vellum 300 years ago. In addition to this, segmenting handwritten Ge'ez numbers and punctuation marks result segmentation error [4].

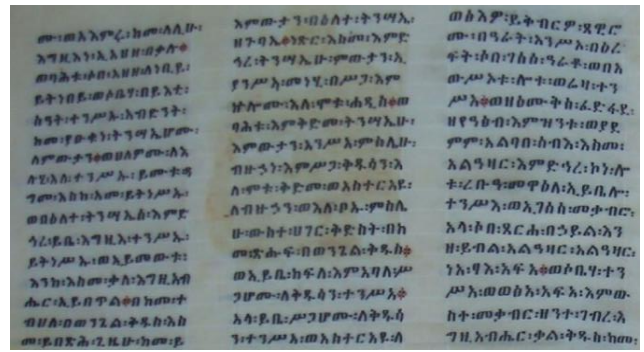

Figure 1: Sample document image taken from Haymanote-Abew (ሃይማኖተ አበው) vellum book

This brings the necessity to develop an OCR system for Ge'ez scripts. As a result, in this study, we developed an OCR system that recognizes Ge'ez scripts which were written on vellum by the applying modified segmentation approach.

## 2. LITERATURE REVIEW

The recognition of handwritten document is considered as one of the most challenging problems in the field of pattern recognition [1]. The main principle in automatic recognition of patterns is first to teach the machine, which classes of patterns that occur and what they look like. In line with this, attempts have been made to develop OCR systems for Ethiopic (Amharic and Ge'ez) scripts.

Million [2], developed an OCR system for computer printed Amharic texts. The performance of the OCR system was evaluated using different test cases written in WashRa, Agafari_Addis Zemen and Rejim and Visual Geez font types. The proposed OCR system was insensitive to font variations.

Abay [5], attempted to integrate noise detection and removal techniques to develop optical character recognizer for real-life Amharic documents. Different preprocessing algorithms, namely adaptive noise removal, global Otsu thresholding, stage-by-stage segmentation, normalization, thinning techniques and neural network classifier methods had been experimented.

Wondwossen [6], attempted to develop an OCR engine for a special type of handwritten Amharic text, traditionally called "Yekum Tsehuf". He showed that one should be very careful not to lose important features of handwritten character during binarization. He used a multilayer feed forward neural network for recognition.

Shiferaw [4], developed an OCR system for Ge'ez scripts written on the vellum. In his research, image binarization, noise removal, thinning, size normalization, line and character segmentation, feature extraction with SVM classifier had been experimented.

In reviewing related papers, the segmentation of Ge'ez numbers and punctuation marks were observed as a main cause for segmentation errors [4]. The errors occurred due to the space between the disconnected strokes in Ge'ez numbers and punctuation marks.

The writing style and shape of a single handwritten Ge'ez number or punctuation mark require space between strokes, which is actually the body of a single Ge'ez number or punctuation mark.

Additionally, older handwritten Ge'ez numbers and punctuation marks are constructed by disconnecting strokes. The space between these strokes generates a segmentation error. Moreover, Ge'ez scripts written on vellum have poor quality with high noise intensity with overlapped, detached, skewed and touched scripts. The implication is that, Ge'ez numbers and punctuation marks segmentation are the actual challenge that motivated this study.

## 3. METHODOLOGY OF THE STUDY

### 3.1 Image acquisition

In this research, the datasets were collected from different older vellum Ge'ez books, namely Haymanote-Abew /ሃይማኖተ አበው/, Abushaker-Bahire Hasab /አቡሻክር-ባህረ ሃሳብ/, Retua Haymanot /ርቱዓ ሃይማኖት/, Zena Eskindir /ዜና እስክንድር/, Aba Haile-Micheal /አባ ሀይለ-ሚካኤል/, Metsihafe Kelimentose /መጽሐፈ ቀሌምንጦስ/.

In order to develop the recognizer, we have used a total of 191 Ge'ez character classes where 170 classes were a collection of primary characters, 17 classes for Ge'ez number symbols and 4 classes for Ge'ez punctuation marks.

### 3.2 Preprocessing

Preprocessing step consists of noise reduction and binarization; adaptive filtering was used for noise reduction and Isodata iterative global thresholding was used for document image binarization.

*A. Noise Removal*

In this study, an adaptive filter with wiener2 MATLAB function was used to remove noise from the document images.

Figure 2: Image after applying adaptive filter

*B. Binarization*

We used Isodata global image thresholding for document image binarization.

Figure 3: Image after applying Isodata thresholding

*3.3 Segmentation*

Line and character segmentation were taken place to segment Ge'ez characters, numbers and punctuation marks. Since the collected image documents were written 300 years ago, scribers used punctuation marks such as neus-neteb /ንዑስ ነጥብ ፡/, netela-serez /ነጠላ ሠረዝ ፣/, neus-serez /ንዑስ ሠረዝ ፤/, sebate-neteb /ስብዐተ ነጥብ ፨/, ebiy-serez /ዐቢይ ሠረዝ ፥/ to fill space between words. Thus, word segmentation is not performed.

**Line segmentation:** We used horizontal projection method to segment lines from the input image. A sample document image line segmentation result is shown in Figure 4.

Figure 4: Line segmentation output

However, the line segmentation algorithm has encountered segmentation errors. This is due to the occurrence of unwanted extra pixels between segmented lines. The line segmentation failure is shown in Figure 5.

(a)
(b)

Figure 5: Line segmentation error (a) The original line (b) Image detached from the original line (error line)

**Character Segmentation:** The modified bounding box projection approach was used for character segmentation. When bounding box projection was used directly to segment Ge'ez numbers and punctuation marks, segmentation errors were encountered. It segments disconnected strokes in their individual bounding boxes, as shown on Figure 6. To perform segmentation by tolerating the gap between disconnected strokes of Ge'ez numbers and punctuation marks, the bounding box projection approach was modified.

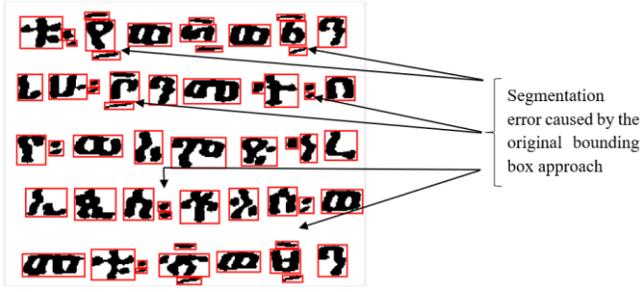

Figure 6: Segmentation errors generated by bounding box projection approach

## Modified bounding box projection algorithm

To resolve the character segmentation error, we have used morphological operations, erosion and dilation with rectangle structuring element along with the bounding box projection method. The algorithm is depicted on Algorithm 1.

### Algorithm 1: Modified Bounding Box Projection

    **Input**: BinaryImage
    **Output**: SegmentedChar
    NewBinaryImage←bwareaopen (BinaryImage, P);

    dilateBW←imdilate(NewBinaryImage, strel(res);
    erodeBW←imerode(dilateBW, strel(res);

    [L Ne]←bwlabel(erodeBW);

    S←regionprops (erodeBW, 'BoundingBox');
    **for** n=1 *to* length(S) **do**
        rectangle ('Position',
        S(n,:).BoundingBox, 'EdgeColor', 'r','LineWidth', 1);
    **end**
    figure;
    increment=1;
    **for** n=1 *to* Ne
     [r, c]←find(L==n);
     SegmentedChar←NewBinaryImage (min(r): max(r), min(c): max(c));
    **end**

### Algorithm description:

We typically choose a structuring element the same size and shape as the objects we want to process in the input image. Rectangle morphological structuring elements were then implemented, given the average *size [m, n], (2-element vector of positive integers)* for the bounding box objects.

We used to fill the gap between disconnected strokes of Ge'ez numbers and punctuation marks before the bounding box encloses the entire characters; via a dilation morphological operation. This operation takes automatic advantage of the decomposition of a rectangular structuring element. Binary image packing was also used to speed up the dilation process.

The morphological erosion operation was performed to erode the binary image with an average *size [m, n]* defined for an array of rectangle structuring element objects.

In Algorithm 1, the binary erosion of the *dilateBW* image by *res*, denoted *dilateBW ⊖ res*, is defined as the set operation:

$$dilateBW \ominus res = \{r | (res_r \subseteq dilateBW)\},$$

Which means, *r* is the set of pixel locations, where the structuring element translated to location *r* overlaps only with foreground pixels in *dilateBW*. Where,

- *dilateBW* – the dilated image and an input for binary erosion operation
- *res* – an object holds an array of rectangle structuring element with *size [m, n]*
- *r* – specific pixel locations in the structuring element

After the erosion operation was performed, 8-connected objects found in the binary image were labeled to return the matrix of L as shown in Algorithm 1. 8-connected objects imply that pixels are connected if their edges or corners touch. Two adjoining pixels are part of the same object if they are both on and are connected along the horizontal, vertical, left-to-right diagonal or right-to-left diagonal direction [8].

By measuring the properties of image regions, bounding boxes are plotted to enclose any grouping of connected black pixels in the eroded image. The size of the smallest bounding box containing the region, returned as a *1-by-(2\*Q)* vector. The first *Q* components are the directions of the minimum corner of the bounding box. The second *Q* components are the size of the box along with each dimension [9].

To return the vectors of indices for the pixels that make up a specific object we labeled the connected components in conjunction with the row and column coordinates of the labeled object. That is denoted by *[r, c]* at Algorithm 1.

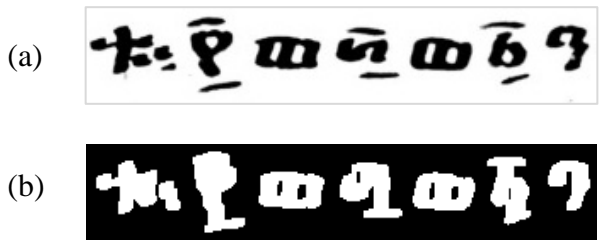

(a)

(b)

Figure 7: Morphological dilation operation (a) before dilation (b) after dilation

As it is shown in Figure 7(b), the space between disconnected strokes of Ge'ez number (፳, ፶, ፷) and punctuation mark (፥) are filled with pixels. The rectangle bounding box encloses these characters after morphological erosion operation is performed. Such pixels filling was not applicable using the bounding box projection algorithm, as it is shown in Figure 8 (a).

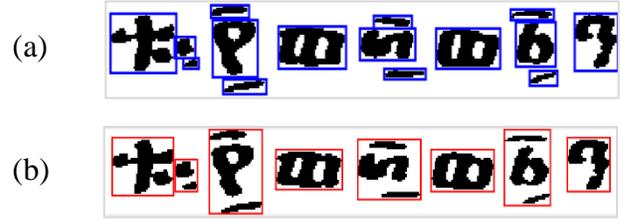

(a)

(b)

Figure 8: Segmentation result (a) Bounding box (b) Modified bounding box

## 3.4 Feature extraction

After the characters were segmented, the next step was extracting features from the segmented images. Figure 9, show a binary matrix representation of Ge'ez character "ω" with 15x15 character dimension. The binary number 1 represents on (black) pixels, where 0 represents off (white) pixels.

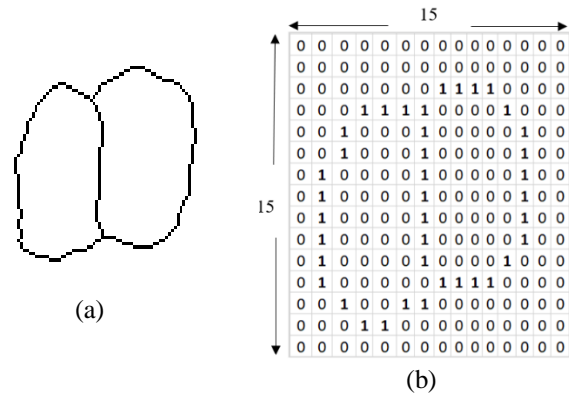

(a)

(b)

Figure 9: Binary matrix representation of Ge'ez character "ω" (a) Thinned character (b) Binary matrix representation

We considered 11 features extraction techniques: Zoning, Moment invariants, Euler number feature, Normalized area of character skeleton, Centroid of image, Eccentricity, Zoning density features, Histogram of Oriented Gradient (HOG), Extent, Connected component analysis and Horizontal projection profile.

At this point, our goal was finding a set of parameters or feature values that define the shape of the underlying character as precisely as possible.

**Global Feature Vector**

The global feature vector was created by appending all the above feature extraction vectors in a single feature vector. There was a 683-dimensional global feature vector for each Ge'ez character, number and punctuation marks. As a result, we got a 683x1528 matrix to train 1528 samples and a 683x382 matrix to test 382 samples.

*3.5 Classification and Mapping*

After the features were extracted the next task was feeding the global feature vector for the classifier. Each Ge'ez character, number and punctuation marks were represented by a certain dimension of feature vectors.

The classification task was done by multiclass classification using SVM classifier. To map the recognized characters into an editable file we used the Unicode standard version of 10.0 for Ethiopic ranges from 1200 to 137F.

### 4. RESULT DISCUSSION

In this research, we used 10 handwritten variations, with different structure, size, and appearance for 191 Ge'ez character classes. A total of 1910 different samples were used to measure the performance of the modified segmentation algorithm. Out of 1910 samples, we used 80% for training and 20% for testing. Analysis of segmentation results scored by the bounded box and the modified bounded segmentation algorithm are depicted in Figure 10.

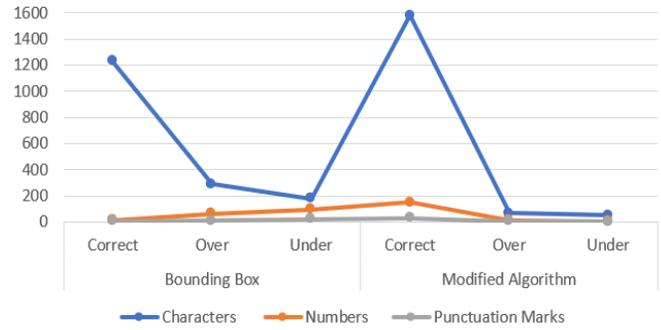

Figure 10: Analysis of segmentation results scored by bounding box and modified algorithm

As shown in Figure 11, high percentage of correctly segmented characters, numbers, punctuation marks scored by bounding box were 72.4%, 7.04% and 12.5%, respectively. Comparatively, 93%, 87.05% and 79.9% correctly segmented characters, numbers and punctuation marks were scored by modified segmentation algorithm, respectively. From the experiments we observed that modified segmentation algorithm registered better performance over the bounding box segmentation algorithm.

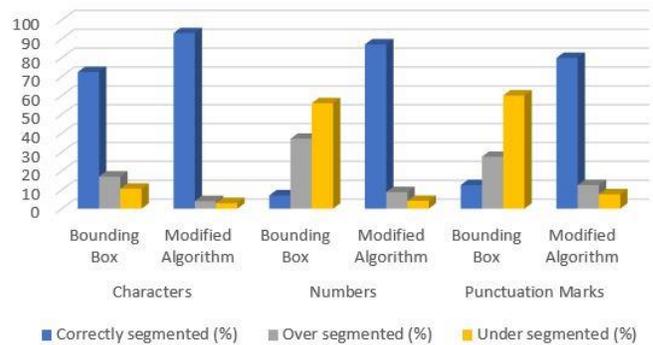

Figure 11: Percentages comparison of segmentation results scored by bounding box and modified algorithm

In this research, linear and polynomial SVM kernel functions scored 79.32% recognition accuracy. However, the performance of the recognizer was challenged by degradation of document images, visual similarity between characters, inconsistent character color, size, style and handwritten variations. Figure 12, show the training and testing accuracy along with misclassification rate scored by SVM linear kernel function.

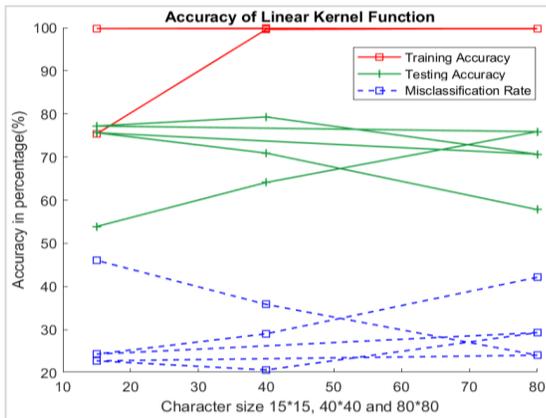

Figure 12: SVM linear kernel function accuracy and misclassification rate

## 5. CONCLUSION AND RECOMMENDATION

Though many attempts had been made; the development of robust OCR engine for Ge'ez scripts is still the most challenging research area.

In this research, we used adaptive filtering technique for noise reduction and Isodata global thresholding for binarization. We showed that the bounding box projection approach was modified, for better accuracy, using erosion and dilation operations. SVM multiclass classifier with the modified bounded box projection technique scored 79.32% recognition accuracy. Overlapping, touched and disconnected characters are segmentation challenges that we did not deal with.